\documentclass[a4paper,UKenglish,cleveref, autoref]{lipics-v2019}

\usepackage{hyperref}
\usepackage{amssymb}
\usepackage{amsmath}
\usepackage{latexsym}
\usepackage{multicol}
\usepackage{booktabs}
\usepackage{multirow}
\usepackage{comment}
\usepackage{amsthm}
\usepackage[ruled,linesnumbered]{algorithm2e}

\usepackage{caption}
\newcommand{\ie}{i.e., }
\newcommand{\eg}{e.g., }

\newcommand{\etal}{et\,al. }

\newcommand{\N}{\mathbb{N}}

\newcommand{\FM}{F\!M}
\newcommand{\BM}{B\!M}
\newcommand{\DM}{D\!M}

\newcommand{\fm}{f\!m}

\newcommand{\F}{\mathscr{F}}

\newcommand{\Rspace} {{\mathbb{R}}}

\makeindex             

\title{Topology-Based Feature Design and Tracking for Multi-Center Cyclones} 

\titlerunning{Topology-Based Feature Design and Tracking for Multi-Center Cyclones}
\author{Wito Engelke}{Scientific Visualization Group, Link\"oping University, Sweden}{wito.engelke@liu.se}{}{}
\author{Talha Bin Masood}{Scientific Visualization Group, Link\"oping University, Sweden}{talha.bin.masood@liu.se}{https://orcid.org/0000-0001-5352-1086}{}
\author{Jakob Beran}{Department of Meteorology (MISU), Stockhom University, Sweden}{jakob.beran@misu.su.se}{}{}
\author{Rodrigo Caballero}{Department of Meteorology (MISU), Stockhom University, Sweden}{rodrigo.caballero@misu.su.se}{https://orcid.org/0000-0002-5507-9209}{}
\author{Ingrid Hotz}{Scientific Visualization Group, Link\"oping University, Sweden}{ingrid.hotz@liu.se}{https://orcid.org/0000-0001-7285-0483}{}

\authorrunning{W. Engelke et al.}

\Copyright{Wito Engelke, Talha Bin Masood, Jakob Beran, Rodrigo Caballero and Ingrid Hotz}

\nolinenumbers 

\EventLongTitle{8th workshop on Topological Methods in Data Analysis and Visualization (TopoInVis 2019)}
\EventShortTitle{TopoInVis 2019}
\EventAcronym{TopoInVis}
\EventYear{2019}
\EventDate{June 17--19, 2019}
\EventLocation{Nyk\"oping, Sweden}
\EventLogo{}
\SeriesVolume{}
\ArticleNo{}

\hideLIPIcs  

\begin{document}
\maketitle

\newcommand{\SubPart}[1]{\emph{\bf #1}}

\abstract{

In this paper, we propose a concept to design, track, and compare application-specific feature definitions expressed as sets of critical points.
Our work has been inspired by the observation that in many applications a large variety of different feature definitions for the same concept are used.
Often, these definitions compete with each other and it is unclear which definition should be used in which context.
A prominent example is the definition of cyclones in climate research.
Despite the differences, frequently these feature definitions can be related to topological concepts. 

In our approach, we provide a cyclone tracking framework that supports interactive feature definition and comparison based on a precomputed tracking graph that stores all extremal points as well as their temporal correspondents.
The framework combines a set of independent building blocks: critical point extraction, critical point tracking, feature definition, and track exploration.
One of the major advantages of such an approach is the flexibility it provides, that is, each block is exchangeable. 
Moreover, it also enables us to perform the most expensive analysis, the construction of a full tracking graph, as a prepossessing step, while keeping the feature definition interactive. 
Different feature definitions can be explored and compared interactively based on this tracking graph.
Features are specified by rules for grouping critical points, while feature tracking corresponds to filtering and querying the full tracking graph by specific requests.
We demonstrate this method for cyclone identification and tracking in the context of climate research.

}

\section{Introduction}
\label{sec:intro}
%
%
Dynamic numerical simulations are prevalent and play a substantial role in understanding physical phenomena.
Usually, such simulations result in feature-rich time-varying multi-fields and appropriate analysis and visualization methods are essential to exploit their full potential.
This entails formal definitions of meaningful features, their algorithmic extraction and tracking, and finally a contextual visualization.
There is a large body of work dealing with these aspects, however, efficient and robust tracking of semantic features at multiple scales is still challenging.
Inspecting existing work, many methods are proposing generic tracking algorithms of topological features as critical points~\cite{Reininghaus2012}, or contours~\cite{Maadasamy2012,Lukasczyk2017}.
While these methods are valuable, they often cannot directly be applied to solve an application-specific tracking problem.
On the other hand, there are methods inspired by applications proposing very specific algorithms for a fixed feature definition, for example for dissipation elements~\cite{Schnorr2018} or tracking of cyclones~\cite{Anil2018}.
Even though these approaches work well in one setting, they often miss the necessary flexibility to be useful in a larger context.
This is partially because physical phenomena can be vague in their description and no commonly accepted mathematical feature definition exists.
Additionally, it is often quite unclear which descriptors work best for which tasks.
A prominent example for such a phenomenon is a cyclone, where new tracking methods are continuously published~\cite{Neu2013}, but still no efficient robust method that is satisfactory for general cyclone tracking exists.
Often, methods have been designed for a specific event, are not generic, and depend on many parameters~\cite{Peixoto1992}.
In this paper, we propose a framework that supports the design and comparison of features defined as sets of critical points, based on robust topological concepts \ie \emph{merge tree} and \emph{Morse complex}.
As an underlying principle, our framework disconnects the extraction and tracking of topological entities from the specific design of features for an application-specific task.
The topological tracking and extraction of critical points results in a large directed graph containing all extrema and their temporal correspondence of the selected scalar fields, as well as a merge tree and its branch decomposition per time-step.
These calculations are performed in a preprocessing step.
Different feature descriptors can then be interactively explored.
Tracking of features is formulated as tracking of groups of extrema and realized as queries to the tracking graph.
As a concrete example, we focus on climate simulation data and the identification as well as tracking of cyclonic systems.
We demonstrate the framework in a meteorological context where understanding the dynamics of weather phenomena is essential to generate reliable predictions of intensity and frequency of extreme weather events in the future.
In Europe, such hazards are mostly associated with extreme extra-tropical cyclones~(ETCs), which are the focus of this paper.
We show how our framework can be applied for efficient identification and tracking of multi-centered systems.
This idea has been successfully applied in \cite{Nilsson2020} for visualization of cyclonic regions.
It formalizes a cyclone identification and tracking method that relies on a few clear principles but is still flexible enough to fulfill the domain scientists' demands.

%
%
The backbone of our method is characterized by a combination of different topological structures, the merge tree~\cite{Carr2003} and the Morse Complex~\cite{Robins2010,Gunther2012a}, facilitating the advantages of both.
The merge tree is well suited for a hierarchical feature definition also supporting multi-centered cyclones, which can be defined as a set of extremal points based on user-specified criteria.
Our tracking algorithm follows a method similar to Reininghaus~\etal\cite{Reininghaus2011,Reininghaus2012}, which is based on the Morse complex.
Parallel implementations are used for both, the merge tree and Morse complex computation~\cite{Acharya2015,Shivashankar2012}.
The tracking itself is inherently local and fast.
\subsection*{Method overview}
\begin{figure}[!ht]
    \centering
    \includegraphics[width=0.99\linewidth]{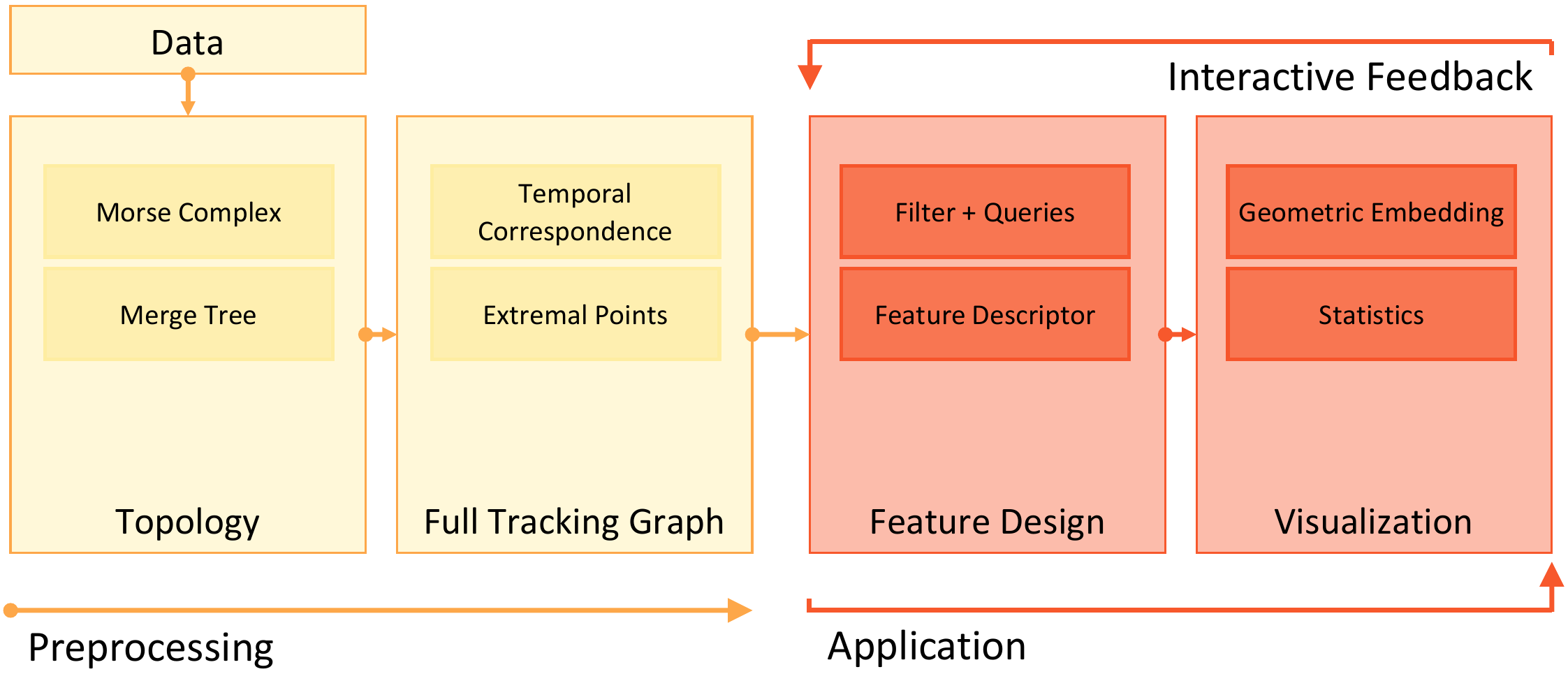}
    \caption{\textbf{Overview:} Different building blocks of our approach. Our method consists of a preprocessing phase, which includes the topological data analysis. The result of this stage is the raw tracking graph containing the temporal correspondence between extrema and the extracted merge tree including its branch decomposition per time-step. In the second building block domain specific knowledge can be used to design features. This step includes the feature descriptor itself, visualizations and computed statistics. This information can also be used to redesign features and reissue tracking graph queries.}
    \label{fig:overview}
\end{figure}
\begin{figure}[h!]
    \centering
    \includegraphics[width=0.99\linewidth]{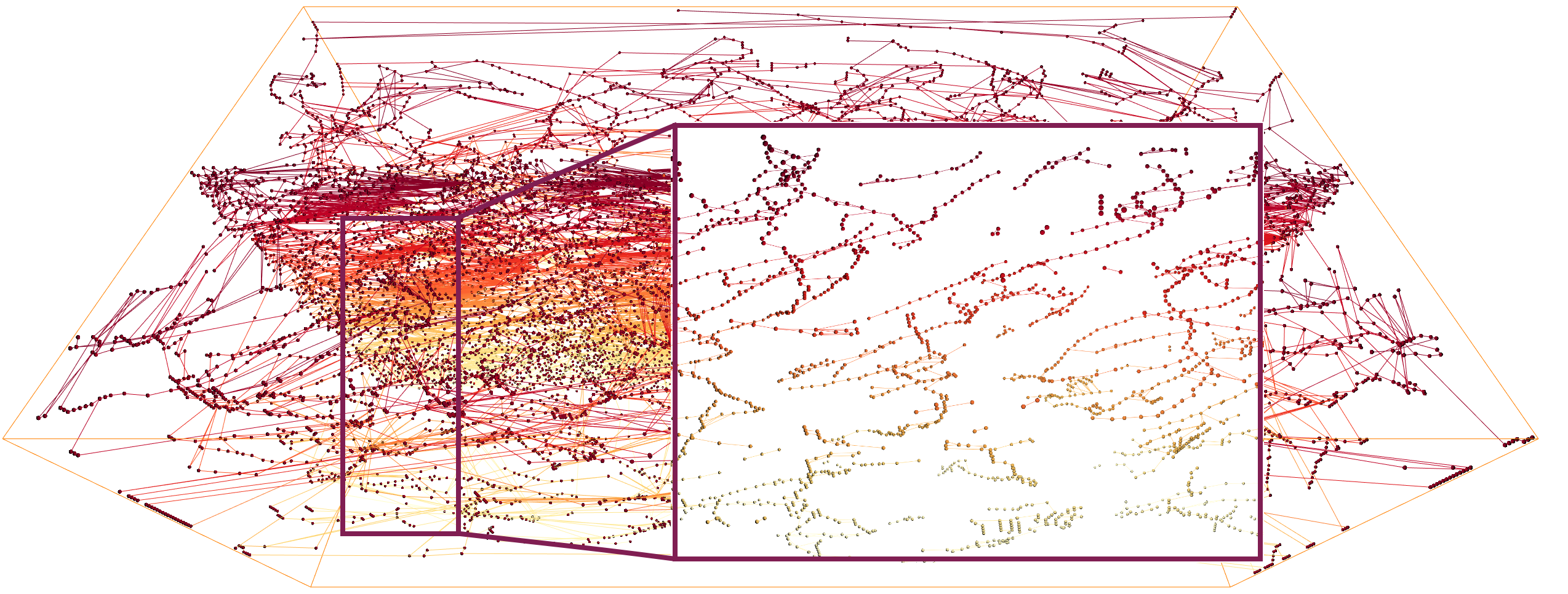}
    \caption{\textbf{Raw Tracking Graph:} Geometric embedding of the complete tracking graph for a two-dimensional time-dependent pressure field given over the globe. The x- and y-axis represent longitude and latitude, whereas the z-direction corresponds to time. The tracking graph contains all minima as vertices and all elements from the forward and backward map as directed edges. This graph itself is independent of any feature definition or filter criterion. The inset shows a filtered subset of the graph. As filter criterion, a combination of edge length and spatial location was used.}
    \label{fig:graph_1}
\end{figure}

In summary, the key aspect of our work is the separation feature definition and tracking.
For both aspects  robust methods from the field of topological data analysis are used.
Fig.~\ref{fig:overview} describes our pipeline in detail.
The tracking approach relies on using the Morse complex to map every extrema in each time-step in both forward and backward directions to construct the raw tracking graph (see Fig.~\ref{fig:graph_1}).
The second building block is an interface for flexible and interchangeable feature definition as a set of critical points obtained from the merge tree including its branch decomposition computed per time-step.
This information is enriched with domain specific criteria and heuristics to form possible feature definitions.

Our main contributions are:
\begin{itemize}
    \item An extrema tracking method based on the Morse complex for time-varying scalar fields.
    \item A concept for flexible feature definitions based on sets of critical points obtained from a merge tree and its branch decomposition.
    \item Application of the above as an example to provide robust definition and tracking of cyclonic systems, possibly containing multiple cyclone centers.
\end{itemize}
%

%
Our feature extraction and tracking is based on two different topological concepts, the merge tree (join or split tree) and the Morse complex where we use the descending/ascending manifolds. Both concepts are briefly described in the following.

\begin{figure}[b]
    \centering
    \subcaptionbox[]{\label{subfig:contourTree_A}}
        {\includegraphics[width=0.40\linewidth]{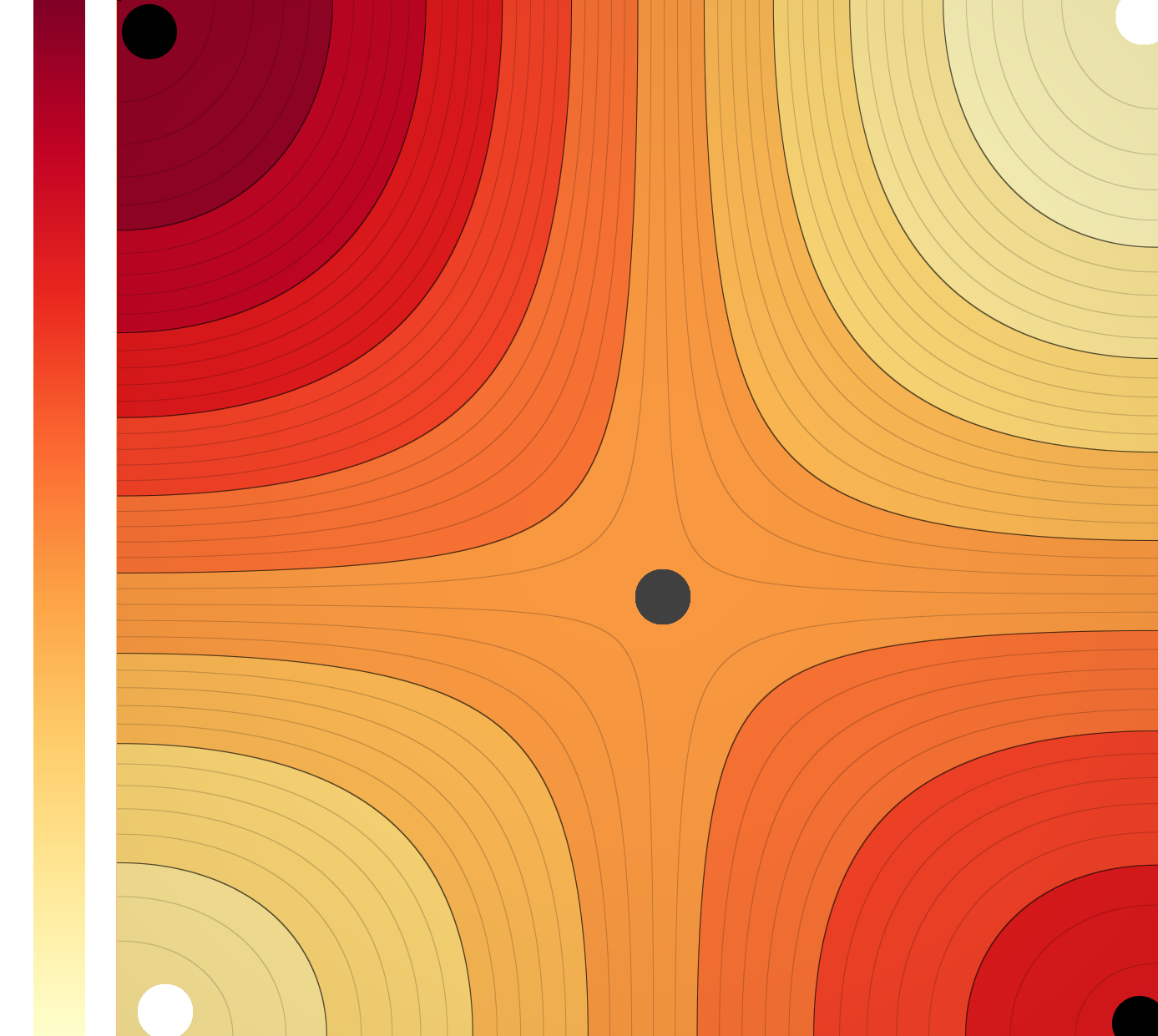}}\hfill%
    \subcaptionbox[]{\label{subfig:contourTree_B}}
        {\includegraphics[width=0.22\linewidth]{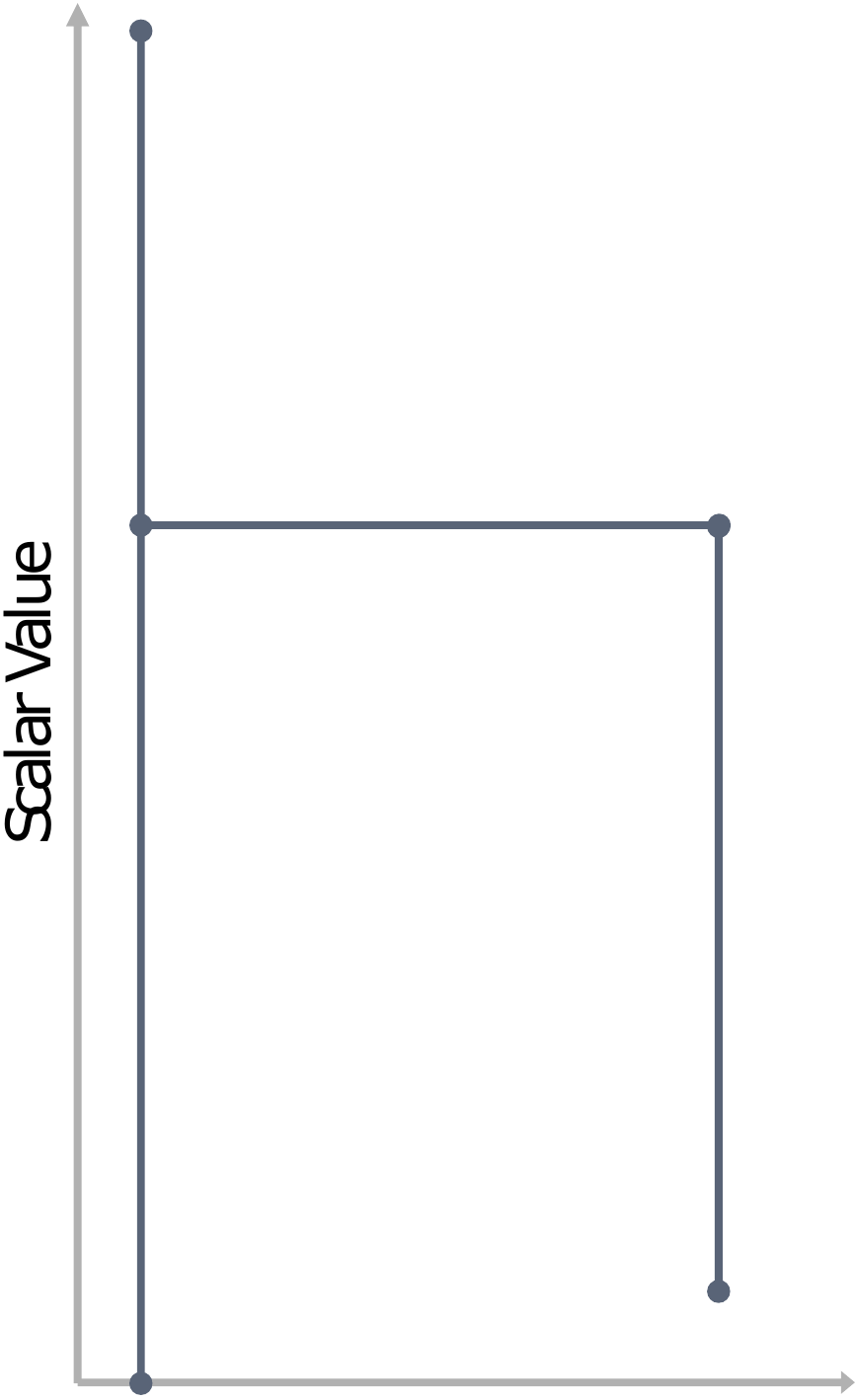}}\hfill%
    \subcaptionbox[]{\label{subfig:contourTree_C}}
        {\includegraphics[width=0.22\linewidth]{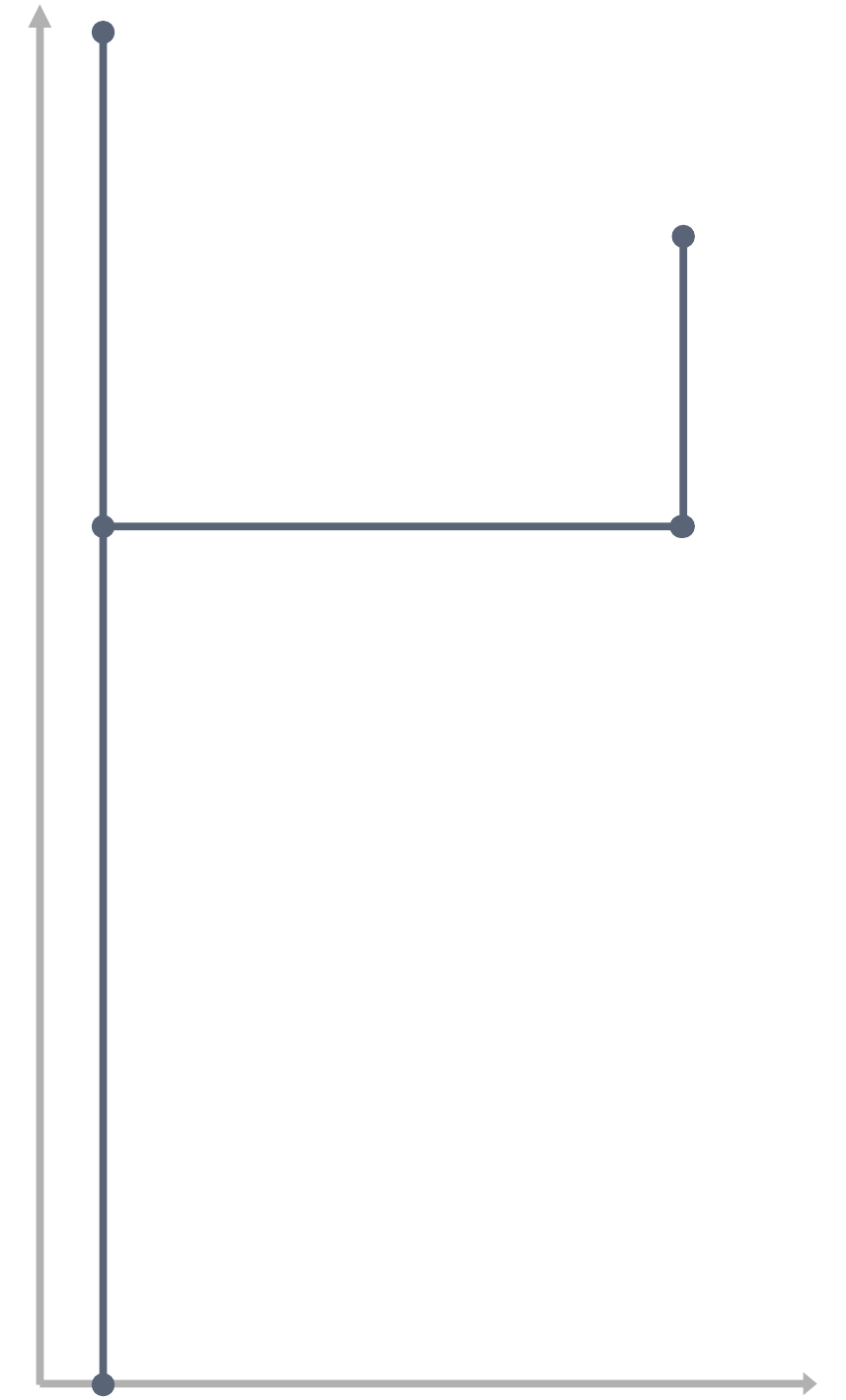}}
        \vspace{-2mm}
    \caption{\textbf{Simple Example:} A scalar field \protect\subref{subfig:contourTree_A} with marked minima (white), maxima (black) and the saddle (gray) on the left and the merge trees in form of a join \protect\subref{subfig:contourTree_B} and split tree \protect\subref{subfig:contourTree_C}.
    }
    \label{fig:mergeTree}
\end{figure}

\paragraph*{Merge Trees}
Features are groups of critical points that are defined by some rules acting on the merge tree of a selected scalar field $f: M \to \Rspace$ defined on a smooth manifold $M$.
Intuitively, a join tree keeps track of topological changes of sub-level sets (or super level-sets in the case of split tree) when changing the level value $a$.
The sub-level sets of $f$ are defined as $M_a := f^{-1}(-\infty, a]$ for some $a \in \Rspace$.
Respectively, super level-sets of $f$ are defined as $M^a := f^{-1}[a,-\infty)$.
Two points $x, y \in M$ are considered \emph{equivalent}, $x \sim y$ if they have the same function value and they belong to the same connected component of the sub-level set $M_a$, respectively the super level-set $M^a$.
A \emph{merge tree} is defined as the quotient space $M / \sim$; it results from identifying points specified by the equivalence relation $\sim$.
The merge tree (resp. split tree) is a graph $G = (V, E)$ rooted at the absolute maximum (resp. absolute minimum), of the field.
Its node-set $V$ consists of local minima and saddle points where the sub-level sets grow together. Its edges $E$ represent the equivalent classes.

Fig.~\ref{fig:mergeTree} illustrates a simple example of join and split trees.
It is constructed by tracking the evolution of the components of $M_a$ as the parameter $a$ is increased (resp. decreased).
Specifically, leaves represent the creation of a component at local extrema, internal nodes represent the merging of components, and the root represents the entire space as a single component.
A merge tree can be embedded in the domain $M$ by visualizing its edges as straight lines that connect spatially-embedded critical points or also be visualized abstractly as a tree~(Fig.~\ref{subfig:contourTree_B} and \protect\subref{subfig:contourTree_C}).

We use a \emph{branch decomposition} of the merge tree that decomposes the tree in a hierarchical structure.
The root branch connects the two global extrema of the scalar field.
All other branches connect a leaf with an interior node corresponding to a saddle point in a hierarchical way such that a persistence-based tree simplification corresponds to removing branches.
For more details, we refer to the work by Pascucci \etal\cite{Pascucci2004}.

\begin{figure}[!hbt]
    \centering
    \subcaptionbox[]{\label{subfig:morseSmale_A}}
        {\includegraphics[width=0.49\linewidth]{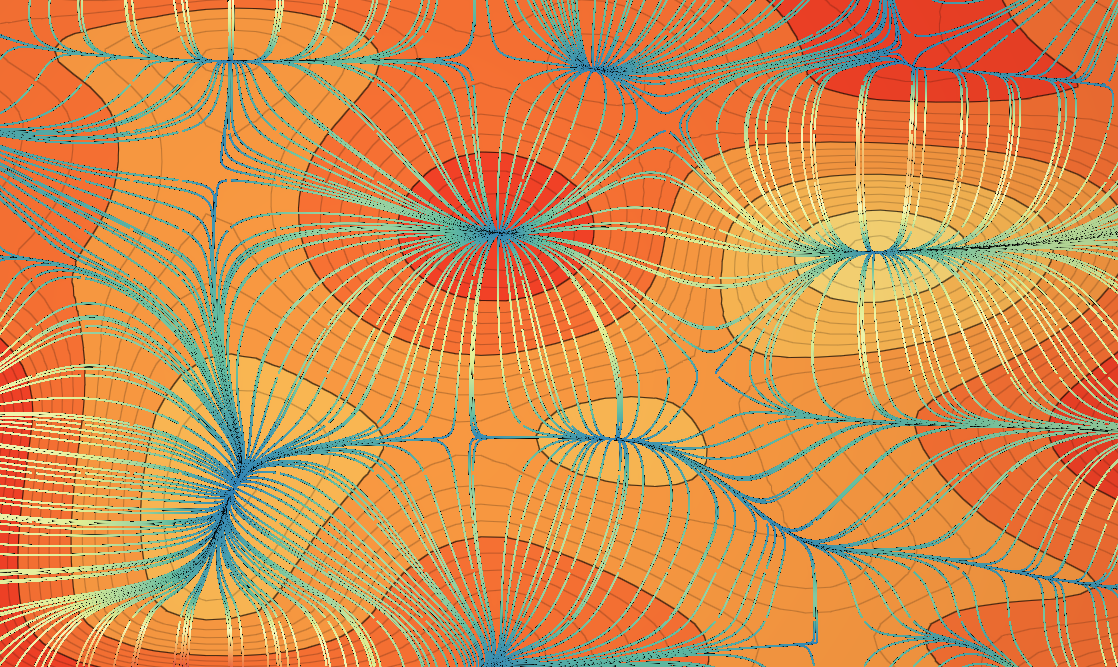}}\hfill%
    \subcaptionbox[]{\label{subfig:morseSmale_B}}
        {\includegraphics[width=0.49\linewidth]{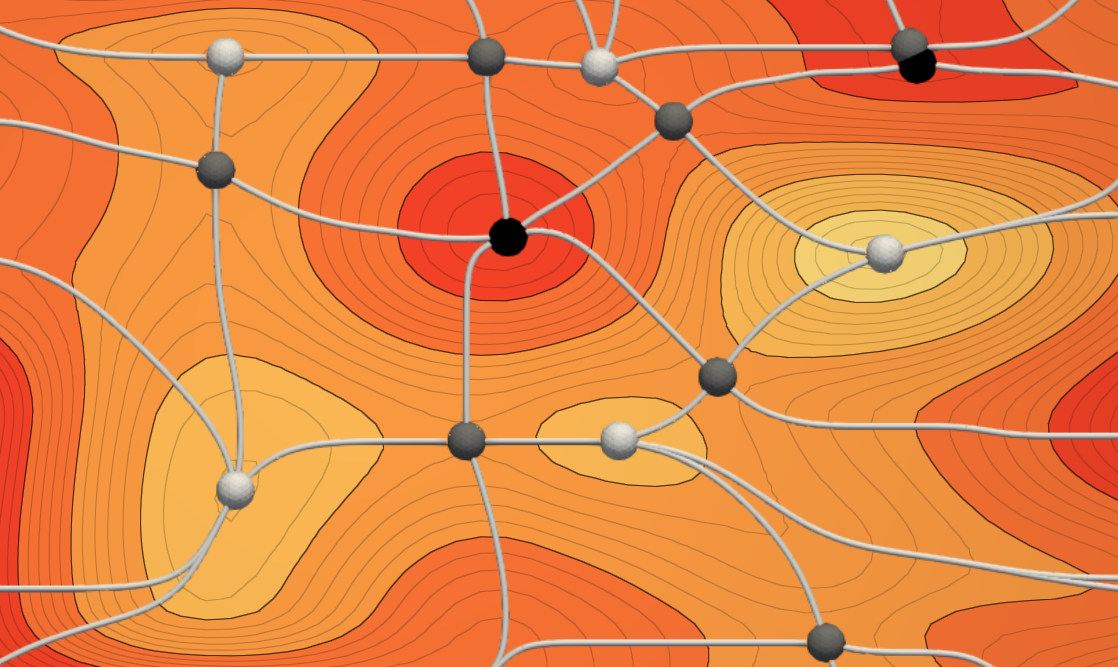}}
    \vspace{-2mm}
    \caption{\textbf{Morse-Smale Complex:} Example of the Morse decomposition of a domain. \protect\subref{subfig:morseSmale_A} shows a visualization of the gradient field derived from a given Morse function. In \protect\subref{subfig:morseSmale_B} the critical points of the function are visualized. White circles indicate locations of minima, black circles maxima, while the saddle points are depicted in gray. Critical points are connected via separatrices. Both images show the same sub-domain.}
    \label{fig:morseSmale}
\end{figure}

\paragraph*{Morse decomposition}
A domain decomposition in ascending (or descending) manifolds forms the basis of our tracking algorithm. These are concepts related to the \emph{Morse-Smale complex} of a function $f$ given over a smooth manifold $M$, in our case of dimension two. A critical point of this function in the domain is a point where all spatial derivatives are equal to zero. In the following, we assume that the Hessian (second order derivatives) at the location of the critical points is non-singular and all critical points have pairwise different function values. Such functions are also called Morse functions.
Based on the function $f$ one can define a decomposition of $M$ that carries geometric and topological information. For this, the gradient of the function $f$, a vector field on $M$, is considered. The critical points of $f$ are the zeros in the gradient field (see Fig.~\ref{subfig:morseSmale_B}).
The Morse-Smale complex then defines a decomposition into regions with uniform gradient flow behavior. This means all gradient lines have the same asymptotic behaviour, emerging from the same minimum and approaching the same maximum. See Fig.~\ref{subfig:morseSmale_A} for an illustration.
These cells can also be interpreted as the intersections of the ascending and descending manifold of critical points. Ascending (resp. descending) manifolds of the critical points are defined as the set of points that flow towards (resp. emerge from) the same critical point.
We are especially interested in the ascending manifolds of maxima and the descending manifolds of minima, which for the two dimensional case are topologically equivalent to open disks.
We use discrete Morse theory for the computation of the cells~\cite{Gyulassy2008, Shivashankar2012}.

\section{Full Tracking Graph Computation}
\label{sec:method_graph}
%
During the preprocessing phase the full tracking graph of a selected scalar field of interest, $f:M\to\mathbb{R}$, is calculated.
This graph provides tracks of all critical points.
It will later serve as a basis for the identification of feature tracks, merging, and splitting events, compare Section~\ref{sec:method_feature}.
Many algorithms that track critical points have been proposed.
In general, the tracking is a two-step process, where critical points are extracted for all time-steps and then a critical point correspondence is established based on some heuristics.
Examples are criteria using distances between critical points or contour overlap~\cite{Lukasczyk2017}. Sohn \etal\cite{Sohn2006} proposed a method to track the entire contour tree using volume overlap.
Treating the time as an additional spatial dimension and assuming linear interpolation between the time-steps, the critical point tracks can be derived using the Reeb graph~\cite{Edelsbrunner2008c, Weber2011}.
In principle, all critical point tracking methods can be used in our framework.
In the current implementation, we decided to follow the concept of combinatorial feature flow fields introduced by Reininghaus \etal\cite{Reininghaus2012}.
After the segmentation of the domain into descending or ascending manifolds, the tracking is a local approach and only requires the evaluation of the next segment a critical point falls into.

Based on the Morse complex as a topologically meaningful partition of the domain, the tracking graph connects all extremal points in the forward and backward temporal directions across consecutive time-steps.
Specifically, we use the descending manifolds for the case of minima tracking and ascending manifolds for maxima tracking.
The tracking method for minima is described in more detail in the following.
Maxima tracking works analogously.
Given an index set $I_t\subset\N$ specifying the minima, let $\{m_i^t,i\in I_t\}$ be the collection of minima of the scalar field $f^t$ in the time-step $t$.
Additionally, let $\DM(m_i^t)$ denote the descending manifold of the minimum $m_i^t$.
We say $m_i^t$ is \emph{forward mapped} to $m_j^{t+1}$ if $m_i^t \in \DM(m_j^{t+1})$. Note that $m_i^t$ and the points in $\DM(m_j^{t+1})$ belong to the same spatial domain, thus checking if $m_i^t$ belongs to  $\DM(m_j^{t+1})$ is a valid operation. Furthermore, in discrete Morse theory, which we use in our implementation, the partition of the mesh vertices into descending manifolds of minima is complete. Combined, the above two conditions ensure that the forward mapping operation is well-defined for all minima $m_i^t$ in time-step $t$, and each minimum $m_i^t$ is forward mapped to a unique minimum $m_j^{t+1}$ in the time-step $t+1$.
Similarly, $m_i^t$ is \emph{backward mapped} to $m_k^{t-1}$ if $m_i^t \in \DM(m_k^{t-1})$.
In other words, we check into which descending manifold a minimum falls, to determine the next position of the minimum forward (and backward) in time.
In this way we define a forward and backward map for time-step $t$ as sets of corresponding minima pairs in forward and backward temporal direction.
\[\FM_t = \{(m_i^t, m_j^{t+1})| i\in I_t, j\in I_{t+1} \text{ and } m_i^t\in \DM(m_j^{t+1} )\}\]
\[\BM_t = \{(m_i^t, m_j^{t-1})| i\in I_t, j\in I_{t-1} \text{ and } m_i^t\in \DM(m_j^{t-1} )\}\]
Please note that every minimum in time-step $t$ is a member of exactly one pair in $\FM_t$ and $\BM_t$.
Combining these maps over all time-steps results in the sets $\FM=\bigcup_t \FM_t$ and $\BM=\bigcup_t \BM_t$.
The two maps together form a directed graph $\mathscr{G} = (\mathscr{E}, \mathscr{V})$.
Here, $\mathscr{V}$ contains all minima $m_i^t$ from the scalar field across all available time-steps.
Additionally, each element of the forward and backward map are represented by one directed edge in $\mathscr{E}$.
For two consecutive time-steps the map corresponds to a $n:m$ mapping between the minima in the respective time-steps.

\paragraph*{Graph Filtering and Queries}
For the construction of the tracking graph all extremal points of the Morse complex are considered, except zero-persistence points.
This avoids problems arising from the unstable geometric embedding of high-persistence critical points over time, compare Fig.~\ref{fig:timesteps}.
This complete tracking graph builds the foundation for further interaction and semantic feature design.
For this purpose, we allow additional node- and edge-based property-vectors.
With this, additional context information, for example, geometric length of an edge and other scalar values can further support the filtering and querying mechanism.
This is important, since the domain partition the graph is based on is complete, and connections violating domain specific requirements may exist.
Filtering can be based on spatial and temporal conditions or rely on edge and node properties.
During graph queries, connected sub-components are extracted, which describe paths of critical points in the spatio-temporal domain.
Here, queries can be formulated such that specific minima or minima groups are used as request input and current filter criteria are respected.
This gives a powerful tool for exploration and analysis.
%

\section{Feature Definition and Tracking of Cyclonic Systems}
\label{sec:method_feature}
%
\paragraph*{Feature Descriptor} As second building block of our method, the feature design plays an integral role.
The feature descriptor is a set of rules that defines the grouping of the extremal points of one or several scalar fields.
Currently, the merge tree is used as an interface to setup this set of rules.
The specific descriptor is application- and task-dependent and typically also depends on a set of thresholds.
Often, the feature descriptor itself is an active research topic and an interactive interface for designing such descriptor is of great benefit.
In this section, we describe the application of this concept to the topic of cyclone tracking.
Relevant scalar fields in this context are pressure and vorticity fields.
There is no generally accepted feature definition or ground truth for how these fields can be used in a feature descriptor.
In fact, the definition of a cyclone is an open scientific question in the field of climate research.
One common aspect, however, is that features are related to extremal points.
Therefore, our method allows for a generic description of features as long as it is based on sets of extremal points.
During the development of our method, where we closely collaborated with the domain scientists, we focused on cyclone tracking in pressure fields characterized by deep minima.
During the course of this collaboration we experimented with different feature definitions.
Using a global pressure threshold can be seen as a first example of a na\"ive cyclone definition, grouping minima falling into the same component.
Due to the seasonal and latitudinal variations in the pressure field, global thresholds are not the most promising concept but could still allow fast inspections of the extracted feature tracks and are therefore of interest.
A combination of rules based on the merge tree's branch decomposition and the notion of persistence provides a more meaningful cyclone descriptor~\cite{Zomorodian2005}.

\begin{figure}[h!]
    \centering
    \subcaptionbox[]{Marked Feature Carriers\label{subfig:branch_a}}
        {\includegraphics[width=0.48\linewidth]{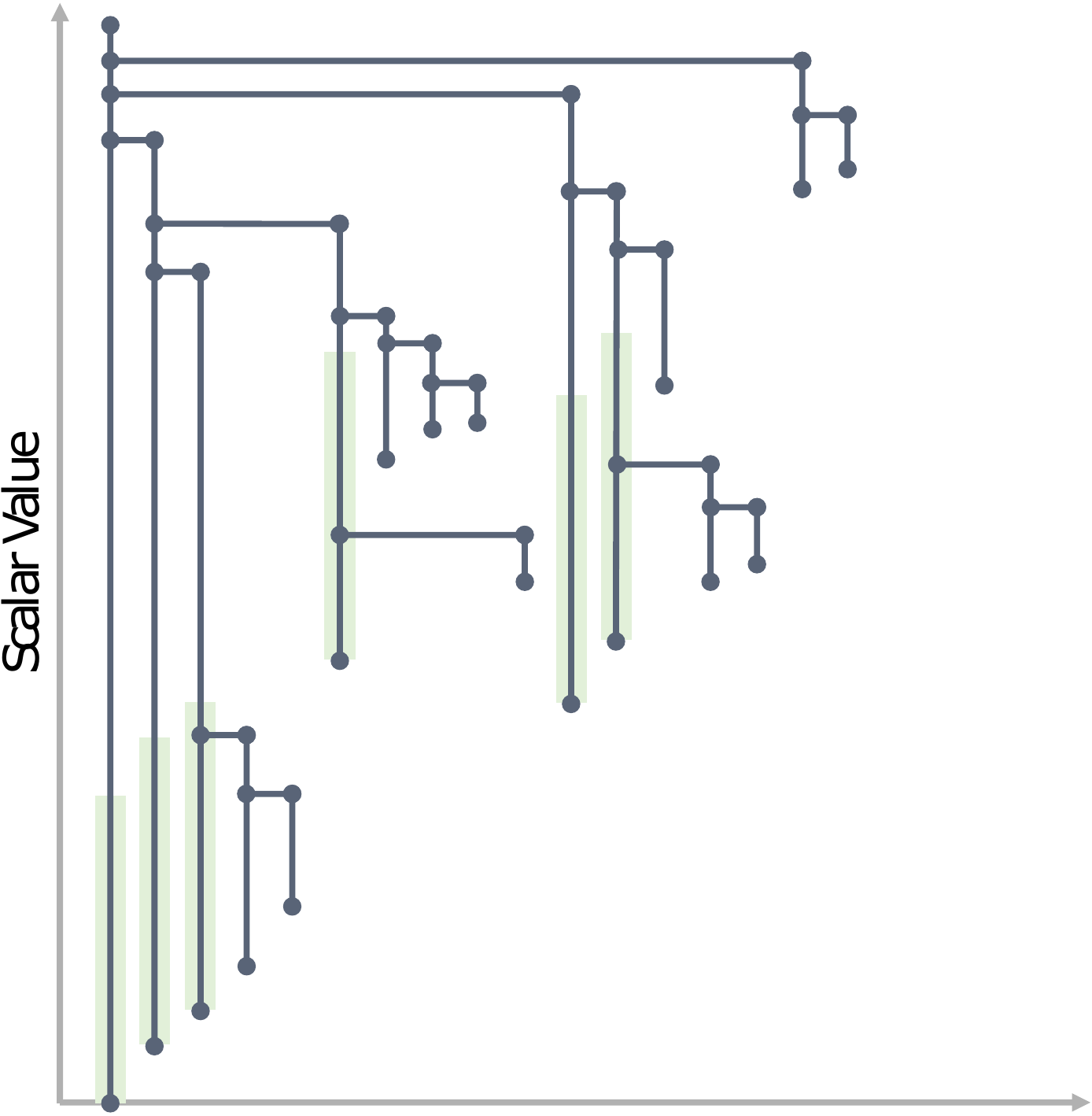}}\hfill
    \subcaptionbox[]{Extracted Features as Sets of EPs\label{subfig:branch_b}}
        {\includegraphics[width=0.48\linewidth]{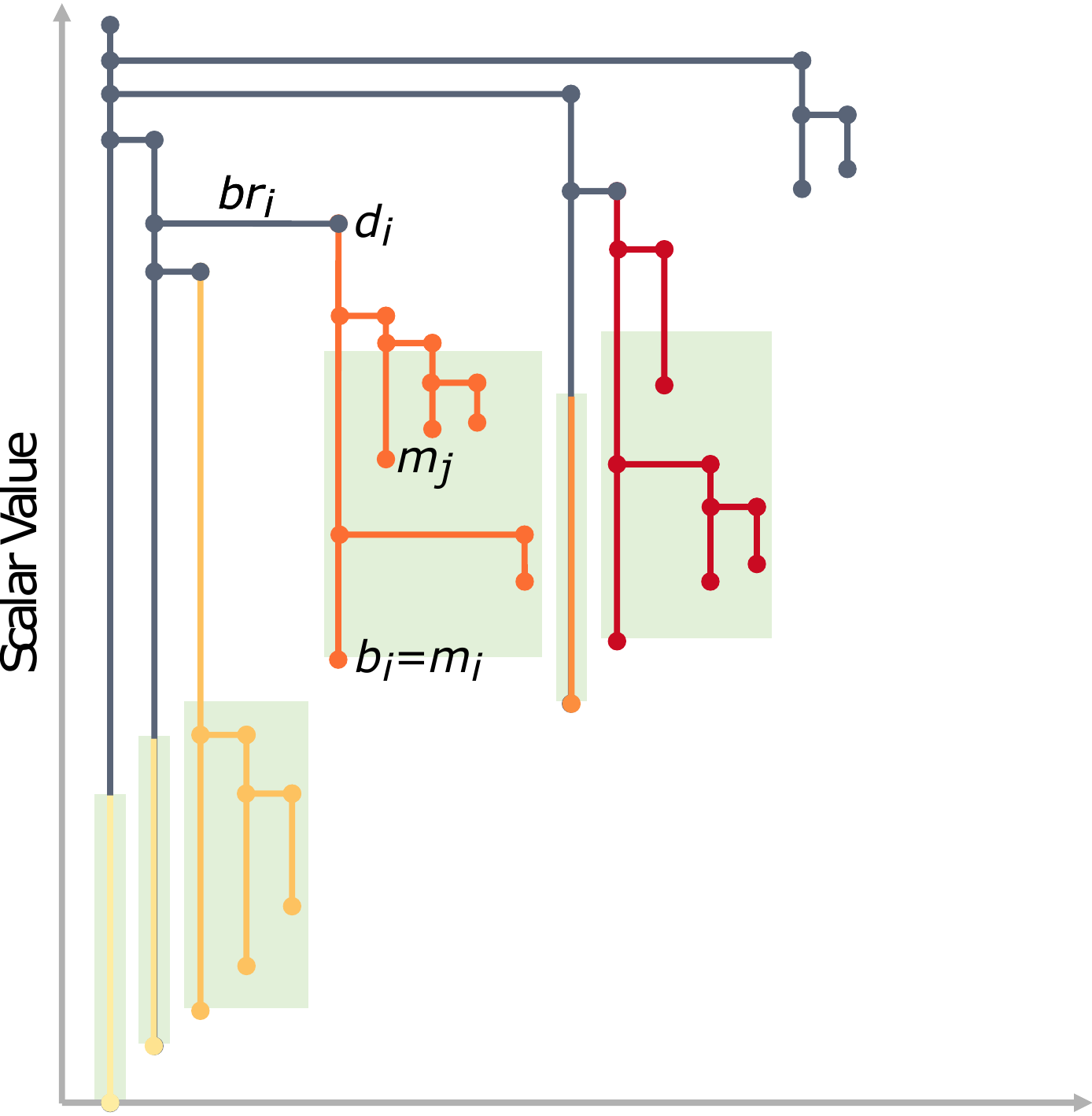}}
    \caption{\textbf{Branch decomposition:} A possible feature definition. \protect\subref{subfig:branch_a} Definition of feature carriers: The persistence threshold $\delta(x) = \delta$ (marked in green) is used to define a cyclonic feature $\F_i$. Only branches with persistence higher than $\delta$ are considered. \protect\subref{subfig:branch_b} Definition of the full feature as set of extremal points. If an extremal point $m_j$ merges into a given feature such that $s(m_j) \leq f(m_i) + \delta$, $(d_j - b_j) < \delta$ and $br_j$ merges into $br_i$, it will be considered as a part of the feature $\F_i$. For each feature, the branch with the highest persistence can be used as the representative. Alternatively, the minimum with the lowest value can be used. For geometric representation, an iso-contour can be used.}
    \label{fig:branches}
\end{figure}

\begin{figure}[h!]
    \centering
    \subcaptionbox[]{$\delta = 2\%$\label{subfig:parameter_02}}
        {\includegraphics[width=0.24\linewidth]{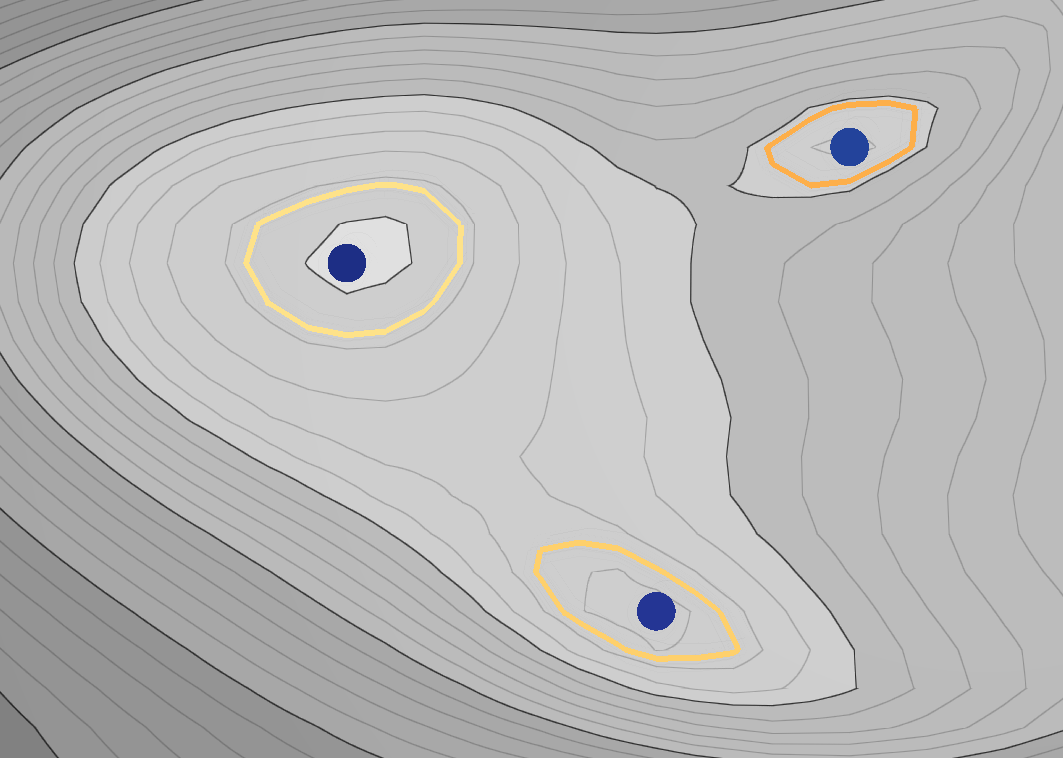}}\hfill%
    \subcaptionbox[]{$\delta = 5\%$\label{subfig:parameter_05}}
        {\includegraphics[width=0.24\linewidth]{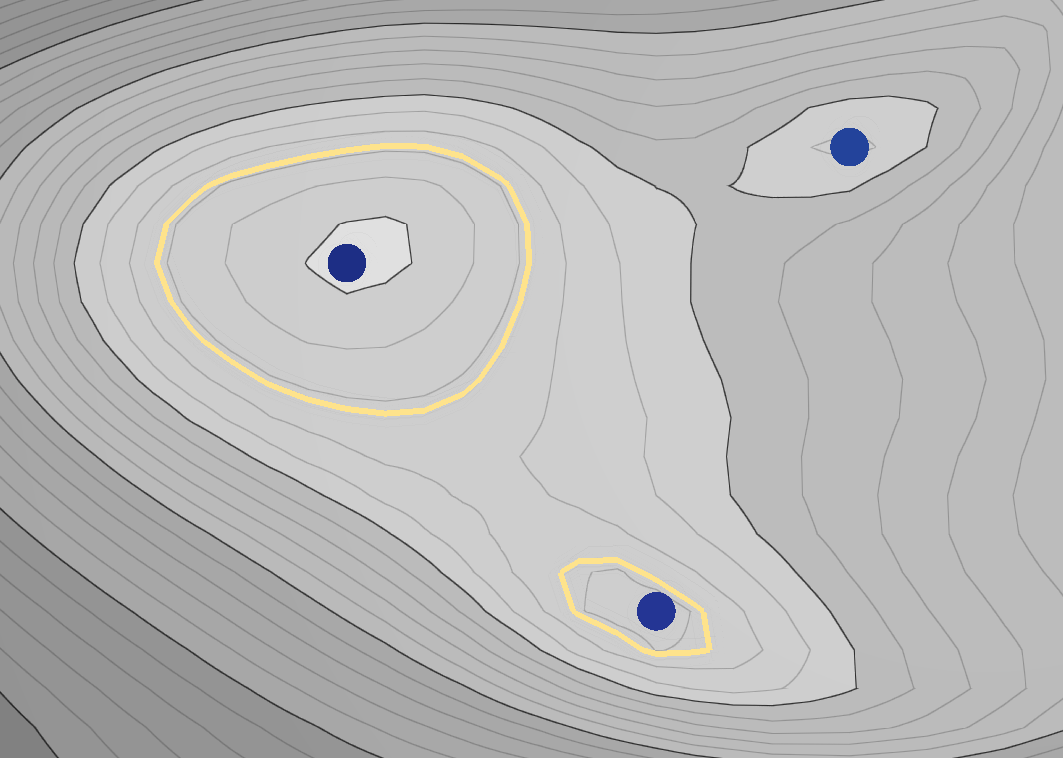}}\hfill%
    \subcaptionbox[]{$\delta = 10\%$\label{subfig:parameter_10}}
        {\includegraphics[width=0.24\linewidth]{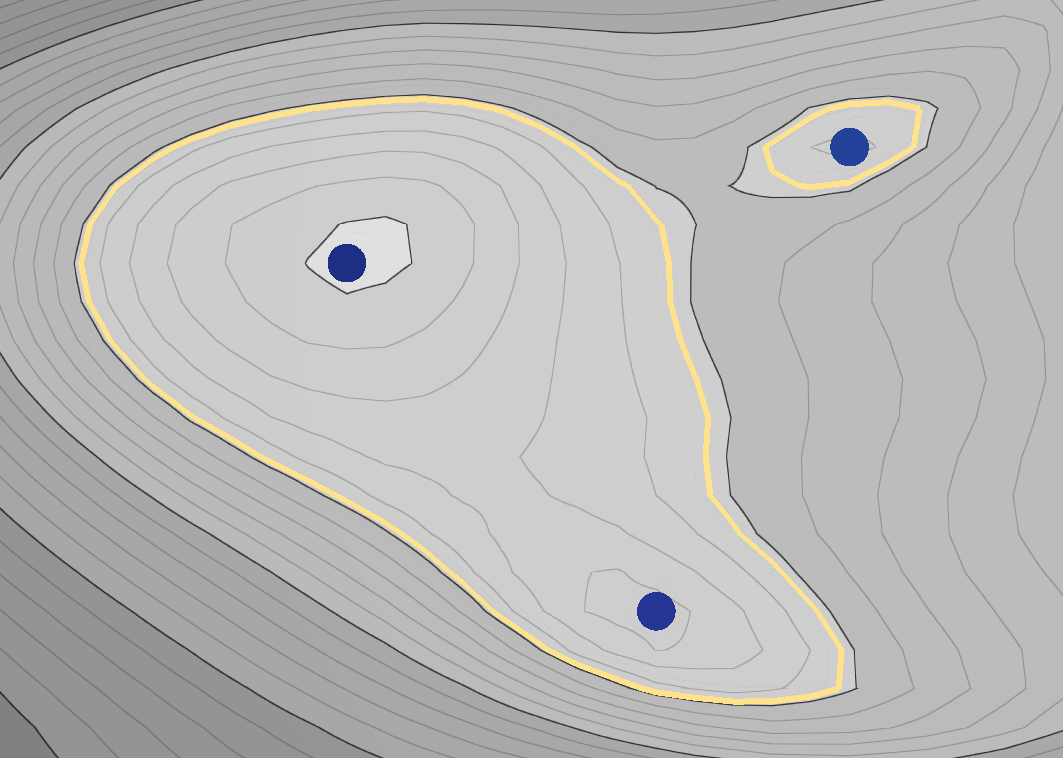}}\hfill%
    \subcaptionbox[]{$\delta = 15\%$\label{subfig:parameter_15}}
        {\includegraphics[width=0.24\linewidth]{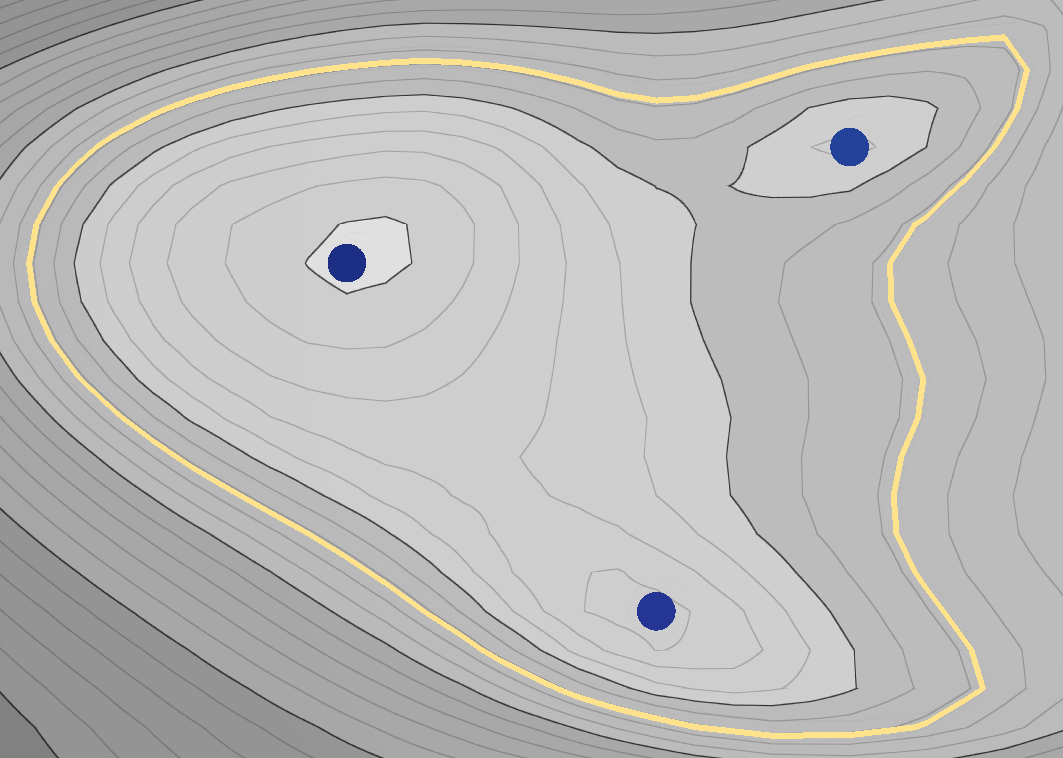}}
    \caption{\textbf{Parameter Influence:} Influence of the parameter $\delta$ in the feature definition. In \protect\subref{subfig:parameter_02} $\delta = 2\%$ of the scalar field range. Each of the three minima spans an individual feature. In \protect\subref{subfig:parameter_05} the value of delta is increased. As a result, the minimum in the top right is not fulfilling the criterion $(d_i - b_i) > \delta$ anymore and therefore does not carry a feature. Further increasing $\delta$ to $10\%$ and $15\%$ leads to a feature consisting of three minima and two separate regions \protect\subref{subfig:parameter_10} and one region \protect\subref{subfig:parameter_15} respectively.}
    \label{fig:parameter}
\end{figure}

At first, one can build on the classic persistence-based filtering approach.
A global persistence threshold can be interpreted as a single rule that groups all critical points of a sub-branch of the merge tree with the depth of the persistence threshold.
However, on its own, this criterion is too rigid to account for the local variations of the pressure field.
An alternative is to define locally varying persistence thresholds which can partially resolve this disadvantage.
All of the aforementioned approaches and their combinations are possible and have their justification.
There are also cases where it is more difficult to formalize a specific intuition.
Here, a manual selection of regions of interest can be part of the setup.
\paragraph*{Local Offset Threshold} 
In the following, we describe one approach that is of specific interest to our collaborators in more detail.
This approach consists of two rules.
The first rule defines a minimum $m_i$ that qualifies as feature carrier for a feature $\F_i$, which is similar to the pure persistence-based approach, see Fig. \ref{fig:branches}(a).
Therefore, a local persistence threshold $\delta$ is defined using the background pressure field.
This threshold can be understood as the `depth of a pressure minimum' to qualify as a cyclone and can depend on the location on earth.
Each branch $br_i=br(b_i, d_i)$ where $b_i=f(m_i)$ is the birth and $d_i$ is the death of this branch with persistence $(d_i - b_i)$ greater than $\delta$ creates a cyclonic feature.
Clearly, the absolute value of the pressure minimum does not matter and instead, the relative depth of the minimum is captured.
The second rule uses three criteria (i)-(iii) to define the set of minima attached to this feature. This results in a complete description of a possible feature definition; where each feature is an extremal point with additional extremal points attached to it, \eg a set of extremal points, see Fig. \ref{subfig:branch_b}.
\[\F_i=\{m_i,m_j| m_j \text{fulfilling criteria (i)-(iii)}\}.\]
with
\[
\begin{array}{rc}
    (i) & f(m_j)\le f(m_i)+\delta\\
    (ii) & (d_j - b_j) < \delta \\
    (iii) & br_j \text{ merges into } br_i
\end{array}
\]
Criterion (i) ensures that the scalar value of $m_j$ is in the interval of the feature defining extremal point itself and $\delta$. Criterion (ii) ensures that the attached extremal point does not span a feature by itself, and criterion (iii) ensures that $m_j$ is part of a child branch of $br_i$.
Each feature $\F_i$ is represented by its master branch $br_i$, which is the branch with the highest persistence within the feature (see Fig. \ref{subfig:branch_a}) or alternatively the lowest minimum $m_i$ in the set.
A possible geometric representation for this feature definition are the components of the iso-contour for the iso-value $s(m_i) + \delta$ which contains at least one minimum of the set $\F_i$ (see Fig. \ref{fig:parameter}).

%
The influence of changing the parameter $\delta$ is demonstrated in Fig.~\ref{fig:parameter}.
It can be seen how the size and the number of extracted features varies.
In general, smaller values lead to smaller features, with respect to the number of extremal points contained in them.
At the same time the total number of features is decreased.

\paragraph*{Feature Tracking}
The feature tracking is based on the full tracking graph.
Considering two features $\F_i^t$ and $\F_j^{t+1}$ in consecutive time-steps, we define the forward tracking score between these features as:
\[
    score(\F_i^t, \F_j^{t+1}) = \sum_{m_i^t \in \F_i^t} \sum_{m_j^{t+1} \in \F_j^{t+1}} w(m_i^t, m_j^{t+1}) \cdot \fm(m_i^t, m_j^{t+1})
\]
with
\[
    \fm(m_i^t, m_j^{t+1}) = \left\{
        \begin{array}{cl}
            1 & \text{if } (m_i^t, m_j^{t+1}) \in \FM \\
            0 & \text{else}
        \end{array}\right.
\]
where $w$ determines the importance of the match between two extremal points in consecutive time-steps.
The weight $w$ can be proportional to the persistence of extremal points, overlap between the descending manifolds or sub-level sets or other reasonable criteria.
Each weight criterion has its advantages and disadvantages. We leave the decision regarding which criteria and weights to use up to the users.
In our implementation, by default, we use $w(m_i^t, m_j^{t+1}) = persistence(m_i^t)$ as the weight function.

Now, for a given feature $\F_i^t$ in the time-step $t$, $score(\F_i^t, \F_j^{t+1})$ can be computed for all features in the time-step $t+1$.
Feature $\F_i^t$ is then mapped to the feature $\F_j^{t+1}$ which has the highest \textit{score} value.
In addition, we also compute the weight of extremal points in $\F_i^t$ which could not be matched to any feature in time-step $t+1$.
If that weight is greater than the maximum matching score, then the feature $\F_i^t$ is not matched to any feature in the next time-step and it dies.

\begin{figure}[!htbp]
    \centering
    \subcaptionbox[]{Corespondence of Extremal Points\label{subfig:feature_track_a}}
        {\includegraphics[width=0.49\linewidth]{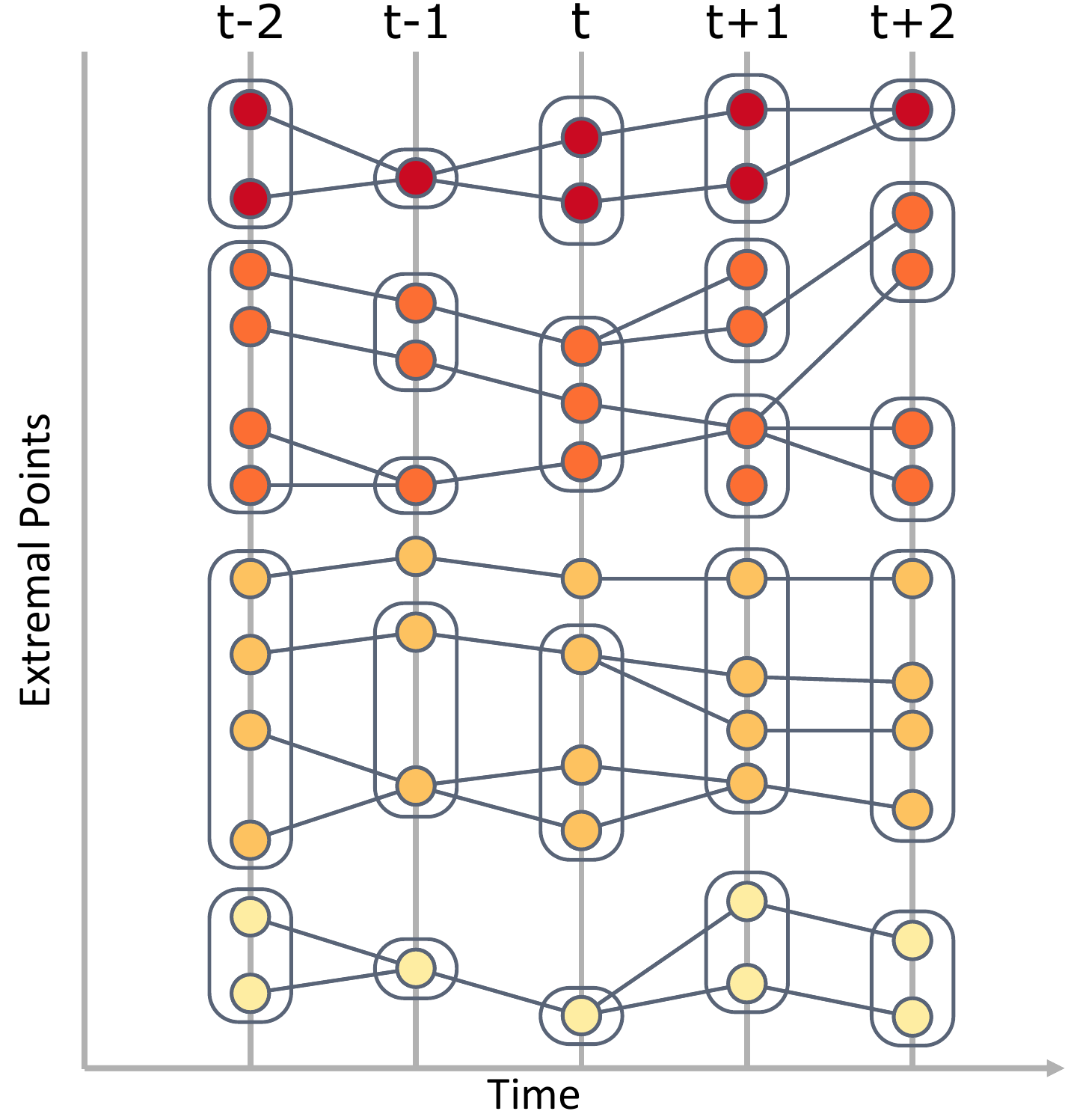}}\hfill%
    \subcaptionbox[]{Feature Tracks\label{subfig:feature_track_b}}
        {\includegraphics[width=0.49\linewidth]{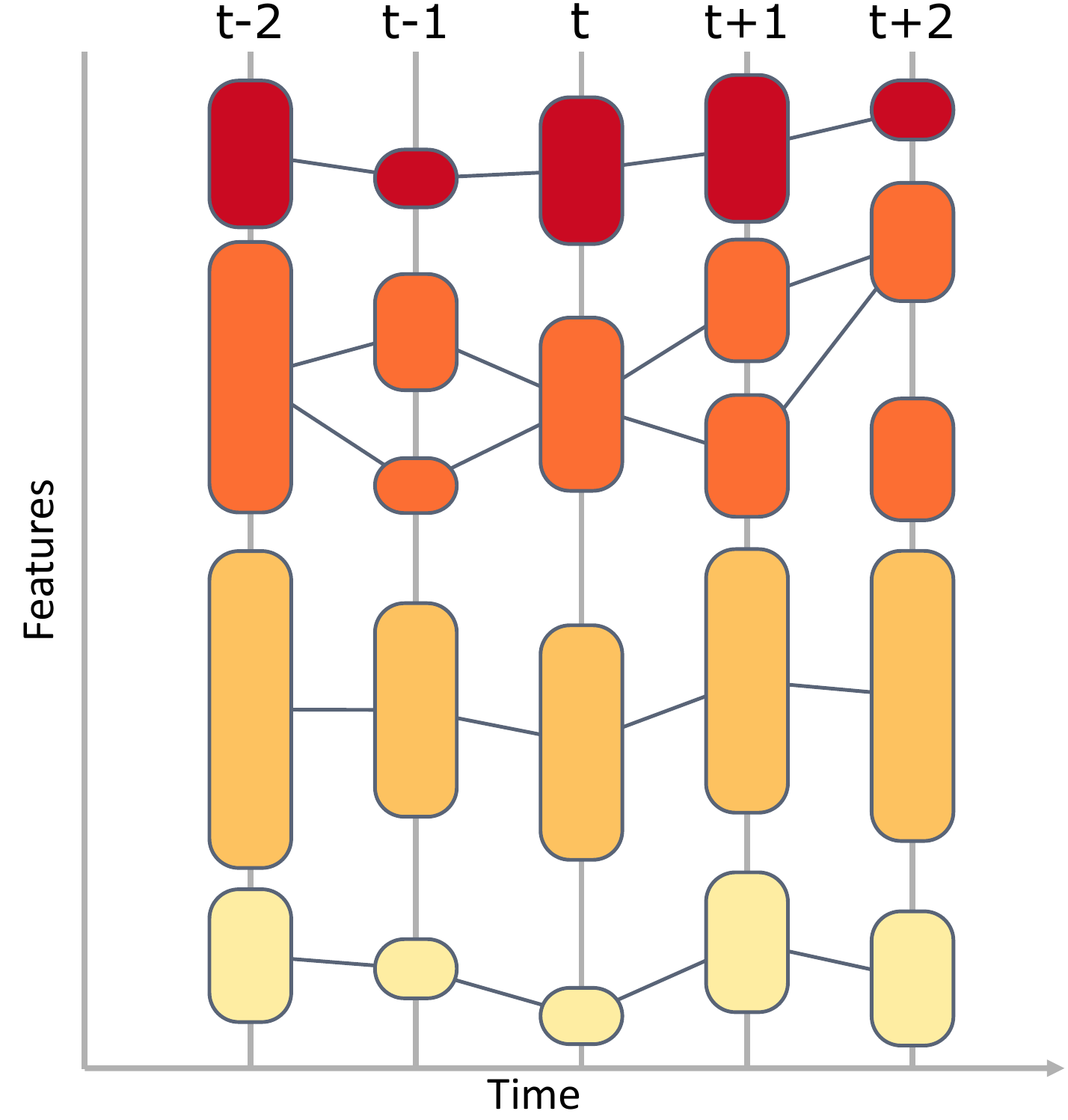}}\hfill%
    \caption{\textbf{Tracking Graph:} Abstract representation of the tracking graph. In both figures each time step is represented as a collection of extremal points (EP) aligned on a vertical axis overlaid by feature grouping these critical points. In the left figure the connections between the critical points illustrate the point-to-point pairs from the sets $\FM$  and $\BM$. The feature tracks live on top of the raw tracking graph for the extremal points, and are highlighted in the right figure. In the feature tracks, unmatched features can exist, supporting the detection of feature birth, death, merge as well as splitting events. However, note that in the raw tracking graph there are no unmatched EPs.}
    \label{fig:feature_track}
\end{figure}

\paragraph*{Tracking Events}
As described in Section~\ref{sec:method_graph}, the raw tracking graph can be filtered and queried.
Recall that this graph only contains mappings between minima and that the used feature description is independent of these mappings.
A minimum can (a) spawn a feature, (b) be part of a feature spawned by a different minimum, or (c) exist without any relation to a feature.
Since this can also change over time, features can be born (birth event) or vanish (death event).
Furthermore, a strongly expressed minimum in one time step spawning its own feature can flatten out and be not pronounced enough to carry its own feature.
If it is close to another strong minimum of similar scalar value the associated features can now merge.
This process and the abstraction between features and minima is illustrated in Fig.~\ref{fig:feature_track}.
With this, features can not only spawn and vanish but also merge and split.

\section{Implementation Details}
\label{sec:implementation}
Our un-optimized prototype implements all algorithmic stages, constructs and stores the raw tracking graph and supports feature descriptor design.
Additionally, various visualization methods for scalar fields, points and edges as needed for the graph and two dimensional plots are implemented within the same framework.
Our implementation uses asynchronous execution and parallel processing where applicable.
Currently, the computation of the Morse complex is the most demanding part and takes about 720~ms for a $320 \times 160$ grid.
During pre-computation of the grid we calculate neighborhood information for the dimension 0 and 1 cells \cite{Gyulassy2008}.
Note that we are not handling dimension 2 cells since we are only interested in descending manifolds.
Computation of the merge tree takes about 170~ms for the same grid size.
Here, we facilitate a union-find data structure as underlying concept.
The given timings are measured on a single CPU workstation running a Xeon E5-1650 v4 with 3.6 GHz.
Rendering is done using the OpenGL API and a combination of a G-Buffer for deferred shading as well as an A-Buffer for order independent transparency.
The current implementation is a proof of concept and we are currently working on the integration of our approach into the \textit{topology toolkit (TTK)} where it also would become open source and freely available to the community.

\section{Case Study}
\label{sec:details}
In this section we present our case study.
Here, we applied our method to the problem of tracking multi-center cyclones in pressure fields.\\

\begin{figure}[hbt!]
    \centering
    \subcaptionbox[]{$t=97$\label{subfig:timesteps_97}}
        {\includegraphics[width=0.32\linewidth]{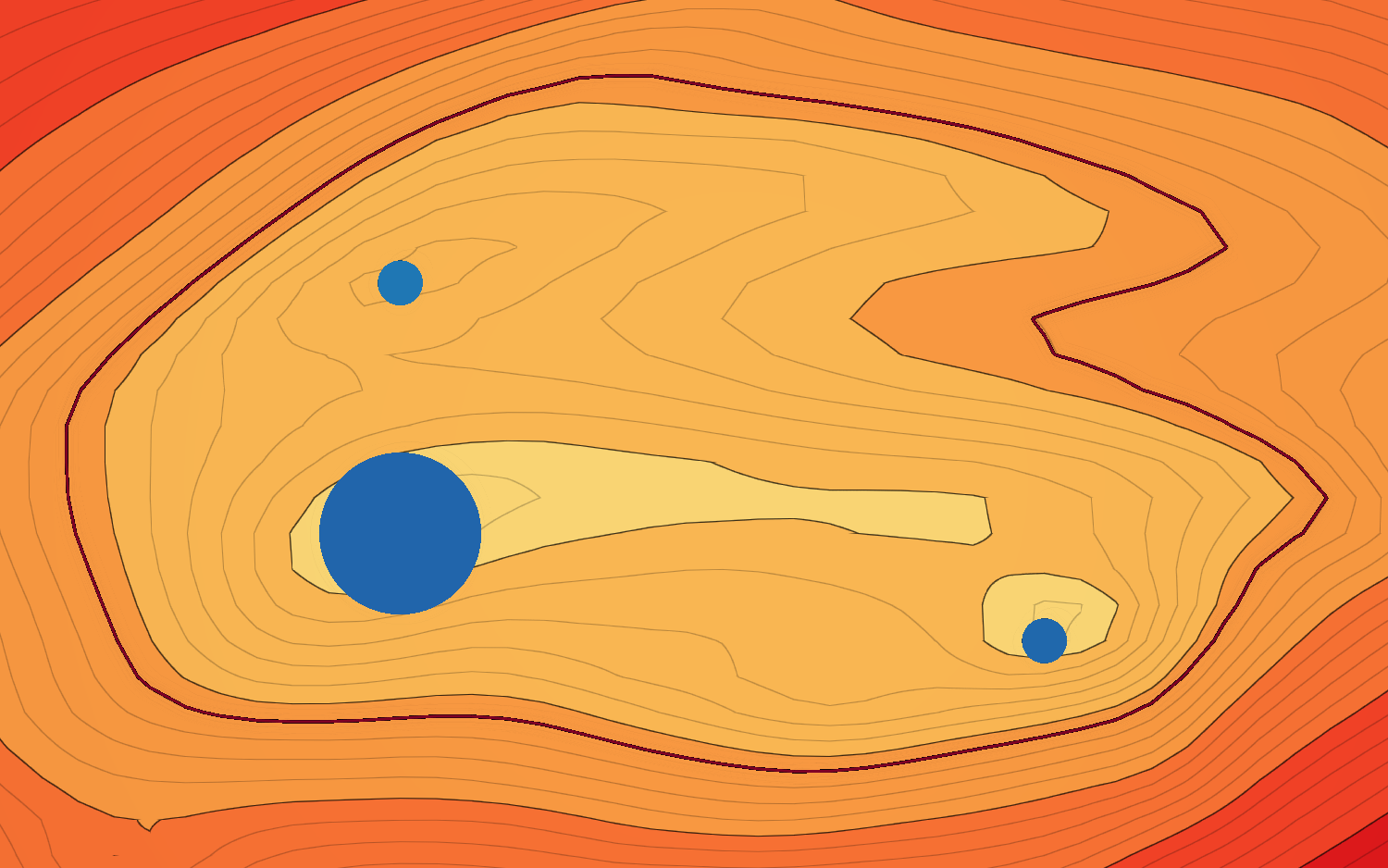}}\hfill%
    \subcaptionbox[]{$t=98$\label{subfig:timesteps_98}}
        {\includegraphics[width=0.32\linewidth]{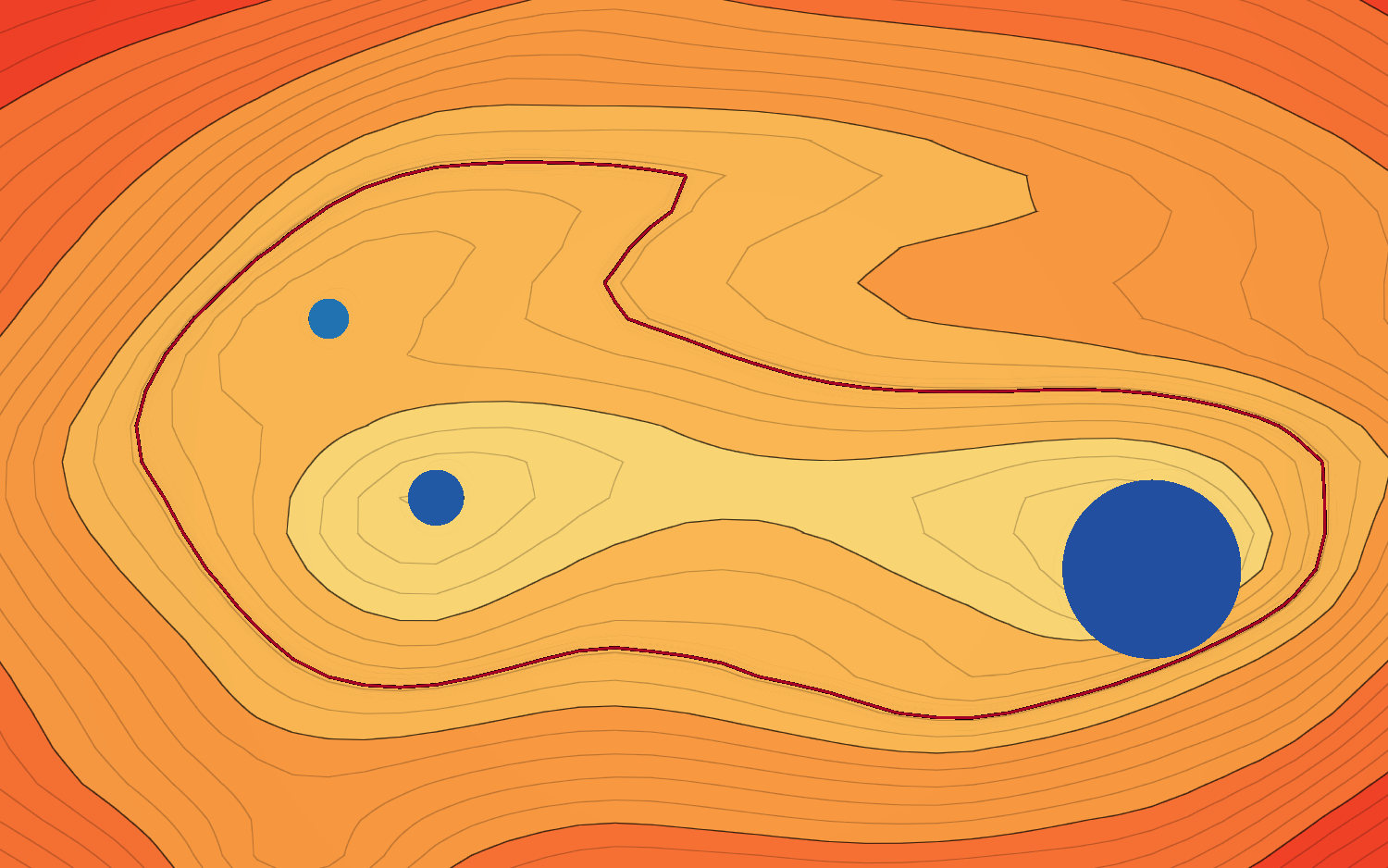}}\hfill%
    \subcaptionbox[]{$t=99$\label{subfig:timesteps_99}}
        {\includegraphics[width=0.32\linewidth]{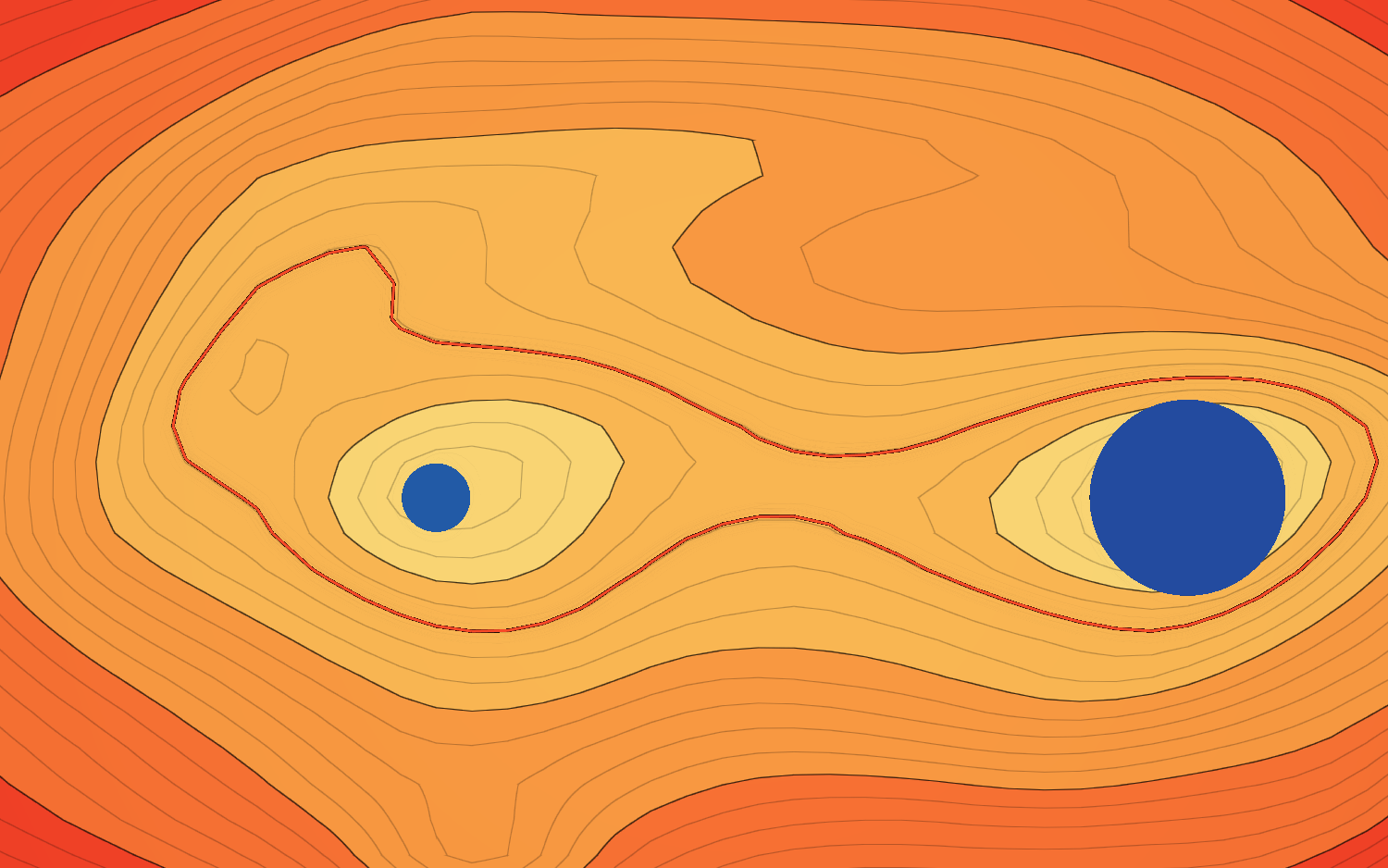}}
    \caption{\textbf{Feature Descriptor:} Three consecutive time-steps of the mean-sea-level pressure field. The spheres indicate the local minima of the field and are scaled by their persistence values. It can be seen that the geometric embedding of the point with maximum persistence is not stable. However, the resulting feature (red iso-line), defined as a combination of the extremal points, is stable.}
    \label{fig:timesteps}
\end{figure}

\noindent
\SubPart{Data Set:} The data set is given as a time-dependent scalar field containing the mean-sea-level~(MSL) pressure of an atmospheric simulation.
The data sets resolution is $320 \times 160$ in longitude and latitude, while the temporal resolution is 120 at 6 hour intervals.
Fig.~\ref{fig:timesteps} shows three consecutive time-steps (\ie time-steps 97, 98, 99) of a subdomain on the northern hemisphere.
The pressure field is rendered by a mapping to a yellow to red color-scale.
Additionally, the visualization is enriched with iso-contours in fixed intervals.\\

\noindent
\SubPart{Feature Descriptor and Tracking Results:} As feature descriptor, we applied a persistence-based metric as described in Section~\ref{sec:method_feature}.
As mentioned, Fig.~\ref{fig:timesteps} shows three consecutive time-steps for a smaller portion of the data-set.
The red iso-line indicate the feature area, whereas the blue circles indicate the location of the local pressure minima.
In addition, the circles are scaled according to the persistence value of these minima.
It can be observed that the geometric embedding of a specific minimum is not stable, meaning the dominant minimum~(\ie the minimum with highest persistence) ``jumps'' from one minimum to another within a single time-step.
Furthermore, a low persistence minimum can appear or vanish within one time-step (see Fig.~\ref{subfig:timesteps_99}).
The used feature descriptor handles both cases as it does not rely on a single critical point.
Therefore, the geometric embedding of the feature itself is stable, even if the geometric embedding of individual minima is not stable.

\begin{figure}[h!]
    \centering
    \subcaptionbox[]{Time Series of Feature Contours\label{subfig:features_all}}
        {\includegraphics[width=0.99\linewidth]{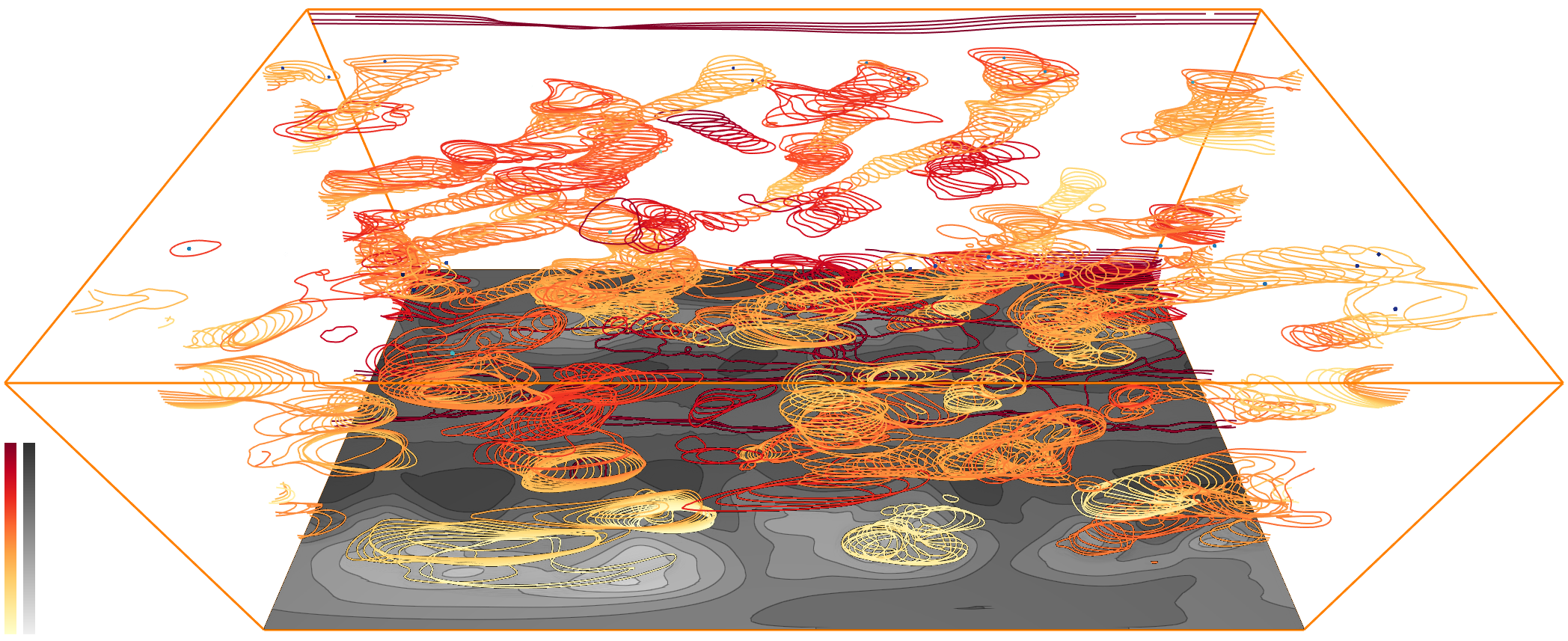}}\hfill%
    \subcaptionbox[]{$t=49$\label{subfig:features_a}}
        {\includegraphics[width=0.32\linewidth]{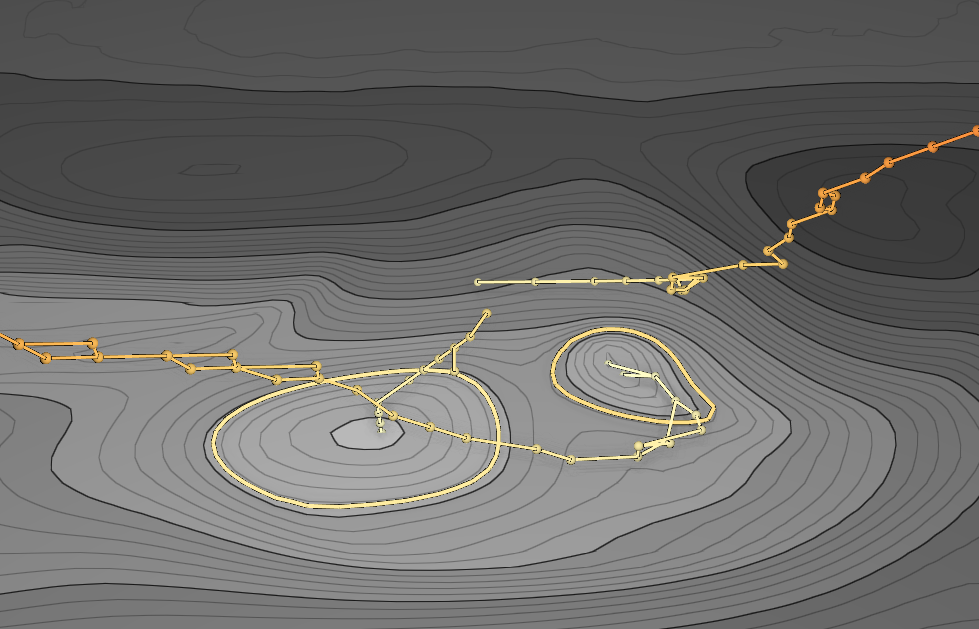}}\hfill%
    \subcaptionbox[]{$t=50$\label{subfig:features_b}}
        {\includegraphics[width=0.32\linewidth]{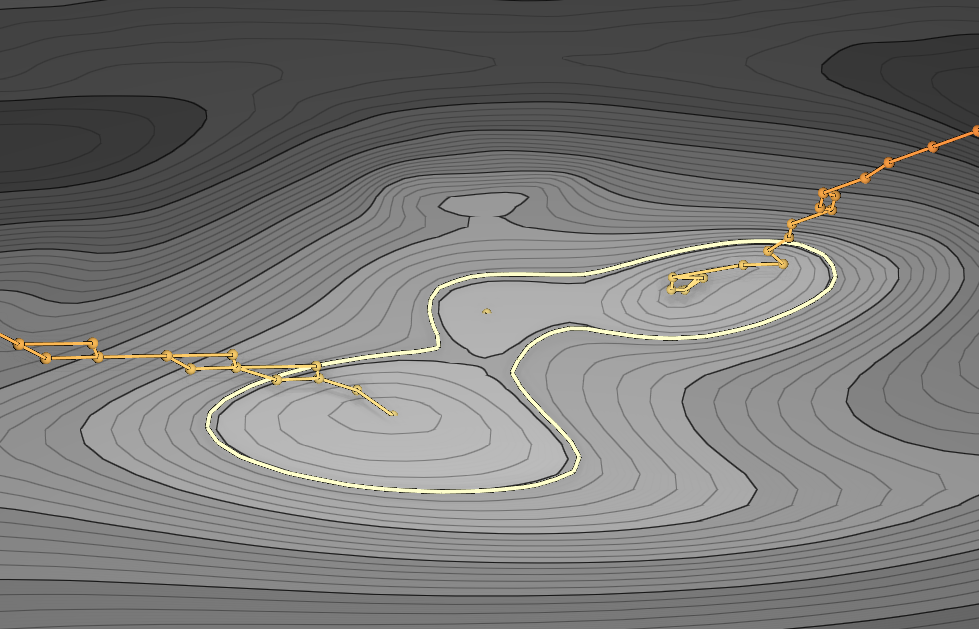}}\hfill%
    \subcaptionbox[]{$t=51$\label{subfig:features_c}}
        {\includegraphics[width=0.32\linewidth]{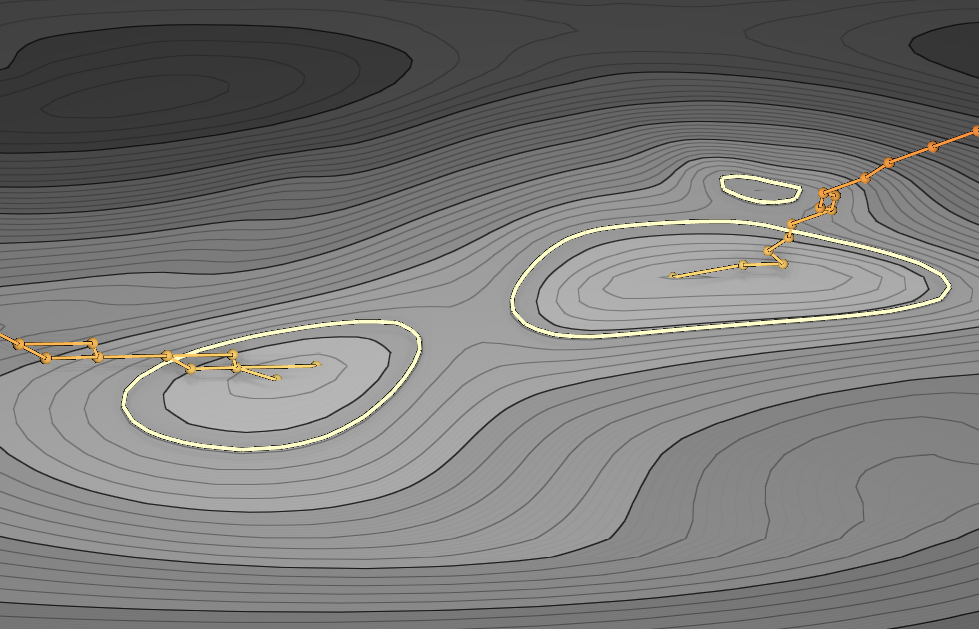}}
    \caption{\textbf{Feature Tracking:} In \protect\subref{subfig:features_all} the iso-contours of all features across all time-steps are shown. The image sequence on the bottom shows the merge of two separate features. In time-step \protect\subref{subfig:features_a} the features are distinct and merge later in time-step \protect\subref{subfig:features_b}. Our tracking graph also contains information about the backward direction which enables us to also detect splitting events. The feature in \protect\subref{subfig:features_b} splits into two separate features after \protect\subref{subfig:features_c}.}
    \label{fig:features2}
\end{figure}

Feature tracking is realized as filtering and querying of our raw tracking graph (see Fig. \ref{fig:graph_1}).
With this, the forward and backward correspondence of features based on minima correspondence can be calculated.
Fig.~\ref{fig:features2} shows a visualization of all features across all time-steps for a specific value of $\delta$.
Furthermore, Fig.~\ref{fig:features2} \protect\subref{subfig:features_a} - \protect\subref{subfig:features_c} show the merge and split of two features over time.

\section{Conclusion and Discussion}
\label{sec:future}
%
We presented a flexible concept to define and track features in scalar data-fields based on topological data analysis as a pre-processing step.
The goal of this approach is not to define yet another cyclone tracking method but to give the application scientists a tool at hand that supports an easy and interactive investigation of changes in their feature definition.
This concept has already been proven to be of value over the course of time when we collaborated with our partners in climate sciences who are experimenting with variations in the definition of cyclones.
The core feature of the method that makes varying cyclone definitions possible is a separation of the topological analysis and the feature configuration. 
The preprocessing step consists of pure topological analysis of the data without any assumptions while following clear rules. 
Thus, it provides a well-defined foundation for an application-specific feature configuration.

A further advantage of this approach is that the computationally expensive steps are separated into a pre-processing step leaving enough headroom to allow for interactive exploration.
Additionally, our current concept can be extended further and adapted to an in situ pipeline. 
Currently, we are using a critical point tracking method that is based on descending or ascending manifolds.
While this tracking gives stable results when following extrema, its accuracy depends on the extraction method of the descending manifolds and can suffer from sub-optimal geometric embedding when using discrete methods.
This is a well-known problem and solutions have been proposed in the literature, e.g.~\cite{Reininghaus2012a} or~\cite{Gyulassy2014b}, which we plan to integrate in the future.
Due to the modular design of our pipeline, the critical point tracking method can be easily exchanged.

In this paper, we have exemplified our concept for the application of cyclone extraction and tracking in pressure fields, but our approach is far more powerful and can be used for other applications as well. 
There are many possibilities for the extension of the work. 
One direction that we want to pursue is to extend the pre-analysis step towards a multi-field analysis. Another interesting idea would be to employ user guided classification of critical points into relevant and non-relevant features in a few selected time steps, and using learning algorithms to automatically extract relevant features in the complete time series.
To improve the feature configuration and feature tracks, there are a lot of ideas in the literature, for example, to consider global optimization criteria for the tracks similar as proposed by Saikia \etal\cite{Saikia2017} or Schnorr \etal\cite{Schnorr2020}.
For this, a better interface for designing rules to be used as feature descriptors is required.
Additionally, we want to extend our prototype in terms of context embedded visualizations and the extraction of statistics, allowing for comparison between different feature descriptors.

\section*{Acknowledgements}
\label{sec:acknowledgements}
This work was supported through a grant from the Swedish Foundation for Strategic Research (SSF, BD15-0082), the SeRC (Swedish e-Science Research Center) and the ELLIIT environment for strategic research in Sweden.
We also thank Jochen Jankowai for his support and many interesting discussions.

\bibliography{literature}

\end{document}